\documentclass[10pt,twocolumn,letterpaper]{article}

\usepackage{ijcb}
\usepackage{times}
\usepackage{epsfig}
\usepackage{graphicx}
\usepackage{amsmath}
\usepackage{amssymb}
\usepackage[norule,symbol,perpage]{footmisc}

\usepackage{cite}
\usepackage{amsmath,amssymb,amsfonts}
\usepackage{algorithmic}
\usepackage{graphicx}
\usepackage{textcomp}
\usepackage{xcolor}
\usepackage{multirow}
\usepackage{colortbl}
\usepackage{amsmath}
\usepackage{amssymb}
\usepackage{cite}
\usepackage[ruled,vlined]{algorithm2e}
\newcommand{\quotes}[1]{``#1''}
\newcommand{\comment}[1]{}
\def\BibTeX{{\rm B\kern-.05em{\sc i\kern-.025em b}\kern-.08em
    T\kern-.1667em\lower.7ex\hbox{E}\kern-.125emX}}

\ijcbfinalcopy % *** Uncomment this line for the final submission

\ifijcbfinal\fi

\makeatletter
\def\ps@IEEEtitlepagestyle{
\def\@oddfoot{\mycopyrightnotice}
\def\@evenfoot{}
}
\makeatother

\begin{document}

%%%%%%%%% TITLE
\title{Attention Aware Wavelet-based Detection of Morphed Face Images}
\author{{Poorya Aghdaie, Baaria Chaudhary, Sobhan Soleymani, Jeremy Dawson, Nasser M. Nasrabadi}\\
{\textit{West Virginia University} }}

% \and

\maketitle
\thispagestyle{empty}

%%%%%%%%% ABSTRACT
\begin{abstract}
  Morphed images have exploited loopholes in the face recognition checkpoints, e.g., Credential Authentication Technology (CAT), used by Transportation Security Administration (TSA), which is a non-trivial security concern. To overcome the risks incurred due to morphed presentations, we propose a wavelet-based morph detection methodology which adopts an end-to-end trainable soft attention mechanism . Our attention-based deep neural network (DNN) focuses on the salient Regions of Interest (ROI) which have the most spatial support for morph detector decision function, i.e, morph class binary softmax output. A retrospective of morph synthesizing procedure aids us to speculate the ROI as regions around facial landmarks , particularly for the case of landmark-based morphing techniques. Moreover, our attention-based DNN is adapted to the wavelet space, where inputs of the network are coarse-to-fine spectral representations, 48 stacked wavelet sub-bands to be exact. We evaluate performance of the proposed framework using three datasets, VISAPP17, LMA, and MorGAN. In addition, as attention maps can be a robust indicator whether a probe image under investigation is genuine or counterfeit, we analyze the estimated attention maps for both a bona fide image and its corresponding morphed image. Finally, we present an ablation study on the efficacy of utilizing attention mechanism for the sake of morph detection. 
\end{abstract}

\begin{figure}[t]
\begin{center}
%\fbox{\rule{0pt}{2in}
\includegraphics[width=.9\linewidth]{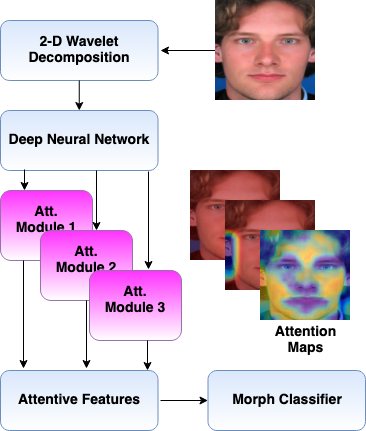} 
\end{center}
   \caption{Our Proposed Deep Attention-based Morph Detector. Wavelet sub-bands of the input image is fed into our DNN during training phase of the network. Three Attention modules, i.e., Att. modules, generate the new attention weighted features, i.e., attentive features, as well as the attention maps. Attentive features are used to detect morphed images. Please note that the attention maps are generated after training the DNN.}
\end{figure}
%%%%%%%%% BODY TEXT
\section{Introduction}

Robust, reliable verification systems are the crucial backbones of biometric document authentication protocols, that are to operate flawlessly. Although image morphing is not a new paradigm, it was first identified as a security concern by Ferrara \etal \cite{ferrara2014magic}, who explained how a criminal can dodge a border control checkpoint using a travel document that was issued with a morphed image. The goal of the face image morphing attack is to synthesize a forged imaged from two composing original images such that the artificially crafted morphed image can be verified against the two original images not only visually, but also in the feature space by a classifier\cite{scherhag2017vulnerability}. Moreover, morphed samples can be labeled as hard positive samples in comparison to negative genuine samples because morphed samples are synthesized to intentionally lie on the negative samples' manifold. Similar to adversarially perturbed data samples that fool classification networks into a wrong predicted class \cite{goodfellow2014explaining, kurakin2016adversarial}, morphed images are crafted to lead a verifier into a false acceptance.

Detecting morphed images has garnered a great deal of attention from the biometrics research community because of its crucial impact on the security protocols \cite{ngan2020face}, especially those used for authenticating travel documents. The vast majority of research efforts has dealt with morphing attacks through either using hand-crafted texture features to find a discriminative hyperplane between the positive (morphed), and negative (genuine) samples \cite{scherhag2019detection, zhang2018face, seibold2018reflection, debiasi2018prnu}, or harvesting those features for learning a deep classifier \cite{debiasi2019detection, scherhag2018morph}. Recently, the visual attention mechanism has taken computer vision community with storm. First introduced in \cite{mnih2014recurrent}, the visual attention mechanism has emerged as a powerful by-product of DNNs, which can boost visual recognition performance on a variety of datasets considerably \cite{jetley2018learn, xu2015show, fukui2019attention, li2018occlusion, wang2017residual}.

In this paper, we present an attention-based DNN in the wavelet domain for detecting morphed samples. To the best of our knowledge, this is the first work which incorporates attention mechanism into a deep morph detector. Our proposed network employs attention to focus on Regions of Interest (ROI) in terms of morph detection, that are specifically landmarks around the eyes and hairline in the landmark-based facial image morphing attacks. 

Wavelet sub-bands of an image represent information with different time-frequency granularity that are adapted to our DNN as input. The soft attention mechanism used in a given layer of our DNN retains spatial regions in the layer's resulting feature maps that represent the discriminative regions, and discard those pixels that are outside the discriminative regions. Fig. 1 shows an overview of our proposed deep attention-based morph detector. We utilize wavelet sub-bands instead of the raw images since we can easily discard frequency contents, sub-bands, which are not discriminative for morph detection such as the low-low (LL) sub-bands. Most importantly, we validate performance of our method through extensive experiments on the three morph datastes: VISAPP17 \cite{makrushin2017automatic}, LMA \cite{damer2018morgan}, and MorGAN \cite{damer2018morgan}. Moreover, estimated attention maps are obtained for both real and morphed images. The contribution of this work are as follows:\begin{itemize}
    \item Incorporating an end-to-end trainable soft attention mechanism into deep morph detector network.
    \item Tailoring wavelet sub-bands for our deep attention-based morph detector. 
    \item Training our deep attention-based network using the three datasets, as well as a combination of all the three datasets, which is coined \quotes{universal} dataset.
\end{itemize}

\section{Related Works}
\subsection{Morph Generation}
Facial morph generation techniques are categorized into two types, i.e., landmark-based morphing \cite{makrushin2017automatic, damer2018morgan, seibold2018accurate, ferrara2014magic}, and GAN-based morphing \cite{damer2018morgan, venkatesh2020can}. In the landmark-based morphing attack, appearance of a resulting morphed image is associated with that of two underlying subject's bona fide face images, while geometric locations of its landmarks are the average of the corresponding landmarks in the two bona fide images \cite{soleymani2021mutual}. By applying Delaunay triangulation on the two bona fide images, corresponding regions on the two facial images are further warped and mixed through alpha blending to synthesize the morphed image. Generative Adversarial Networks (GANs) are also employed for synthesizing morphed images. In \cite{damer2018morgan}, morphed images are generated using a GAN which incorporates an encoder in its generator to model latent space. In addition, morphed images can be generated using StyleGAN \cite{karras2019style, abdal2019image2stylegan}.    
\subsection{Morph Detection}
Different texture descriptors such as Local Binary Patterns (LBP), Histogram of Gradients (HOG), Speeded Up Robust Features (SURF), and Scale-Invariant Feature Transform (SIFT) are utilized for detecting morphed images \cite{debiasi2019detection, 7791169, scherhag2018morph, 8272742}. Discrepancies in  locations of facial landmarks in a morphed image, and a live capture of the corresponding bona fide can be exploited for morph detection \cite{scherhag2018detecting}. Convolutional Neural Networks (CNN), as well as deep embedding features have represented promising performance in morph detection \cite{ferrara2019face, raja2017transferable, aghdaie2021detection, chaudhary2021differential}. In \cite{debiasi2018prnu}, spectral behaviour of Photo Response Non-Uniformity (PRNU) is studied to detect morphed images. The resulting noise artifact in the face morphing pipeline can be adopted for morph detection. In particular, residual noise is an established indicator for morph detection \cite{venkatesh2019morphed, venkatesh2020detecting}. 
\subsection{Attention Mechanism}
Attention mechanism has been widely used for the visual recognition tasks such as image caption generation and visual question answering (VQA) \cite{xu2015show, jetley2018learn}. Two dominant categories of the attention mechanism are the soft deterministic attention and the hard stochastic attention \cite{xu2015show}. The soft attention can either be adopted in a post-hoc manner, or it can be trained along with a DNN using back-propagation \cite{jetley2018learn}. The hard attention mechanism is trained using a method called REINFORCE \cite{williams1992simple}. Attentive recurrent neural networks (RNNs) \cite{mnih2014recurrent} are another variant of networks where expolits attention mechanism to amplify ROI and suppress background clutter. 

\begin{figure*}[t]
\begin{center}
%\fbox{\rule{0pt}{2in}
\includegraphics[width=.9\linewidth]{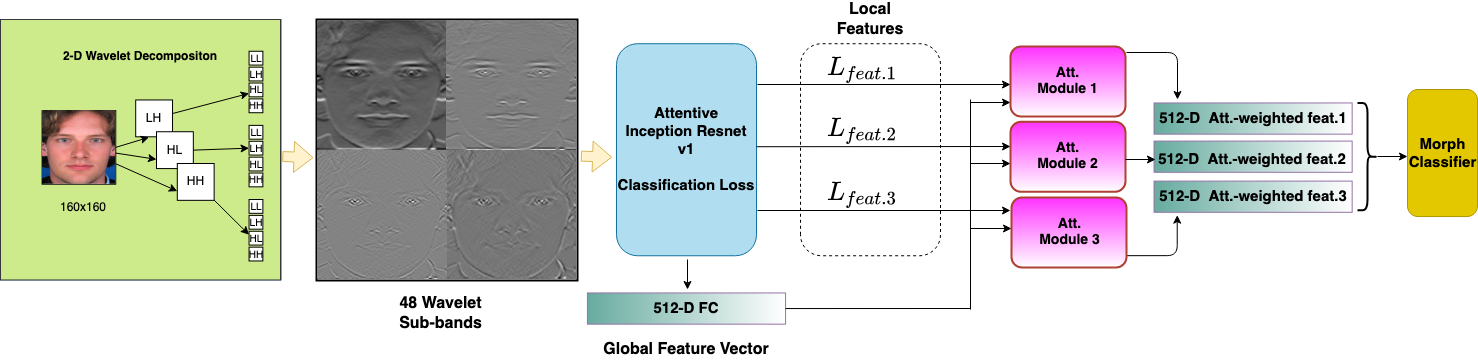} 
\end{center}
   \caption{Our deep attention-based morph detector. The input images are initially decomposed into 48 uniform wavelet sub-bands, which are fed into our morph detector. Attention modules are placed at three convolutional layers, namely $\mathcal{L}_1$,  $\mathcal{L}_2$, and $\mathcal{L}_3$. The $L_{feat.1}$, $L_{feat.2}$, and $L_{feat.3}$ represent the local features vectors of layers $\mathcal{L}_1$,  $\mathcal{L}_2$, and $\mathcal{L}_3$, respectively. The 512-D attention weighted local features in a layer, shown by 512-D Att. weighted feat., are obtained using the local features of the layer, and the 512-D FC global feature vector. The three resulting attention weighted features are concatenated to form our new attended features.}
\end{figure*}

\section{Our Framework}
Our attention-based morph detector is displayed in Fig. 2. Based on Fig. 2, the input images are initially decomposed into 48 uniform wavelet sub-bands that are further stacked channel-wise and then passed to our morph detector. Our morph detector leverages three attention modules at three different convolutional layers, denoted by $\mathcal{L}_1$,  $\mathcal{L}_2$, and $\mathcal{L}_3$. The local feature vectors resulting from the three convolutional layers $\mathcal{L}_1$,  $\mathcal{L}_2$, and $\mathcal{L}_3$ are denoted by $L_{feat.1}$, $L_{feat.2}$, and $L_{feat.3}$, respectively. The attention weighted local features for a given convolutional layer are obtained using the layer's local features, and the global feature vector resulting from the 512-D fully connected (FC) layer in our network, e.g., the first 512-D attention weighted local features in the $\mathcal{L}_1$ layer, shown by 512-D Att. weighted feat.1, are obtained using the local features of $\mathcal{L}_1$, that is to say $L_{feat.1}$, and the 512-D FC global feature vector. The three resulting attention weighted features are concatenated and passed into a new FC layer with $512 \times 3$ neurons.    
\subsection{Uniform Wavelet Decomposition}
Most artifacts due to facial image morphing techniques lie within the high frequency spectrum, and using wavelet decomposition allows us to cherry-pick the desired wavelet sub-bands by discarding the low-frequency sub-bands. Therefore, using specific wavelet sub-bands instead of the original image is highly justified in our study. We apply three-level undecimated 2-D wavelet decomposition on both bona fide and morphed images. Analyzing the wavelet sub-bands of a bona fide and its corresponding morphed image justifies considering the high frequency spectra for the task of morph detection. In other words, we discard the low-low (LL) wavelet sub-band after first level of decomposition, and we keep the low-high (LH), high-low (HL), high-high (HH) for the second and third levels of decomposition. In total, 48 wavelet sub-bands are stacked channel-wise, which are utilized as the input to our attention-based morph detector. Decomposing an RGB image into 48 wavelet sub-bands leads to decoupled spectra, focusing on the frequency contents that are discriminative in terms of distinguishing between bona fide and morphed images.
\subsection{Integrating Attention-Weighted Features}
To distinguish between bona fide and morphed images, we adopt the end-to-end trainable soft attention mechanism introduced in \cite{jetley2018learn}. This soft attention mechanism is differentiable with respect to the network parameters. We show that our attention-based network can meticulously focus on the regions that contribute the most to detecting morphed images. We insert three attention modules at three different convolutional layers $\mathcal{L}_1$,  $\mathcal{L}_2$, and $\mathcal{L}_3$ in our DNN. Therefore, as presented in Fig. 2, instead of a single global feature vector, that is the 512-D fully connected layer (FC) output, we concatenate the three attention-weighted local feature vectors at three different convolutional layers to accomplish the classification task. These attention maps at each convolutional layer reveals the importance of each spatial location in the layers' feature maps. 

Suppose that a spatial local feature vector in the location $i \in \{1,2,..,n\}$ in the convolutional layer $\mathcal{L}_k$, $1\leq k\leq 3$, is shown by $\boldsymbol{\ell}_{i}^{\mathcal{L}_k}$. As presented in Fig. 2, $L_{feat.k} = \{ \boldsymbol{\ell}_{1}^{\mathcal{L}_k}, \boldsymbol{\ell}_{2}^{\mathcal{L}_k}, ...,  \boldsymbol{\ell}_{n}^{\mathcal{L}_k} \}$. The compatibility score for each spatial location, $i$, represents the importance of that pixel for detecting morphed images. The compatibility score for local feature vector $\boldsymbol{\ell}_{i}^{\mathcal{L}_k}$ is given as:
\begin{equation}
{c}_{i}^{\mathcal{L}_k} = \langle \boldsymbol{\ell}_{i}^{\mathcal{L}_k} , \boldsymbol g \rangle, i \in \{1,2,..,n\},
\end{equation}
where $\boldsymbol g$ designates the global feature vector, that is the 512-D output of the fully connected layer and $\langle .,.\rangle$ represents the inner product. 
We further normalize the computed compatibility scores in a given convolutional layer $\mathcal{L}_k$ using the softmax normalization, which is given as:
\begin{equation}
{a}_{i}^{\mathcal{L}_k} = \frac{\exp({c}_{i}^{\mathcal{L}_k})}{\Sigma_{i=1}^{i=n} \exp({c}_{i}^{\mathcal{L}_k})}, i \in \{1,2,..,n\}.
\end{equation}
A linear combination of the local feature vectors $\boldsymbol{\ell}_{i}^{\mathcal{L}_k}$ and the attention weights ${a}_{i}^{\mathcal{L}_k}$ yields the attentive local descriptor for the given convolutional layer $\mathcal{L}_k$. The global feature vector, i.e., attention-weighted feature vector, can be written as:
\begin{equation}
\boldsymbol {g}_{a}^{\mathcal{L}_k} = \Sigma_{i=1}^{i=n} {a}_{i}^{\mathcal{L}_k} \boldsymbol{\ell}_{i}^{\mathcal{L}_k}.
\end{equation}
We concatenate the estimated attention weighted local features at three different convolutional layers which are fed into a FC layer having size of $512 \times 3$ followed by a 2-neuron FC layer, which generates the binary logits for detecting morphed images.

\begin{table}[t]
\caption[Table caption text]{Performance of single morph detection: D-EER\%, BPCER@APCER=5\%, and BPCER@APCER=10\%.} 
\small
\begin{center}
\addtolength{\tabcolsep}{-0pt}
\begin{tabular}{ccccc}
\hline
Dataset&Algorithm&D-EER&5\%&10\%\\ 
\hline
\multirow{6}{*}{\rotatebox[origin=c]{90}{VISAPP17}}&BSIF+SVM~ \cite{kannala2012bsif}&                      16.51 &35.61&26.79\\
                                &SIFT+SVM~ \cite{lowe1999object}&     38.59 &82.40&75.60\\
                                &LBP+SVM~ \cite{ojala1994performance}&  38.00 &77.10 & 67.90\\
                                &SURF+SVM~ \cite{bay2006surf}&  30.45 &84.70 & 69.40\\
                                &RGB+DNN~ \cite{szegedy2017inception}&  1.76 &0.588 & 0.58\\
                                &\textbf{Ours}&                                 {\bf0.00} &{\bf0.00}&{\bf0.00}\\

% &\multirow{5}{*}{\rotatebox[origin=c]{90}{Universal}}&BSIF+SVM~ \cite{kannala2012bsif}&                      23.74 &51.42&38.67\\
%                                 &&SIFT+SVM~ \cite{lowe1999object}&     37.21 &87.45&76.71\\
%                                 &&LBP+SVM~ \cite{ojala1994performance}&  38.80 &91.36 & 83.40\\
%                                 &&SURF+SVM~ \cite{bay2006surf}&  36.00 &75.50 & 65.76\\
%                                 &&Ours&                                 {\bf3.55} &{\bf2.69}&{\bf1.27}\\\cline{2-6}                                       
\hline

\multirow{6}{*}{\rotatebox[origin=c]{90}{LMA}}&BSIF+SVM~ \cite{kannala2012bsif}&                      33.05 &78.34&62.86\\
                                &SIFT+SVM~ \cite{lowe1999object}&     33.30 &83.40&72.00\\
                                &LBP+SVM~ \cite{ojala1994performance}&  28.00 &58.60 & 51.40\\
                                &SURF+SVM~ \cite{bay2006surf}&  37.40 &79.50 & 70.00\\
                                &RGB+DNN~ \cite{szegedy2017inception}&  9.10 &15.18 & 7.49\\
                                &\textbf{Ours}&                                 
                                {\bf 8.71} &{\bf 17.86}&{\bf 6.52}\\

\hline

\multirow{6}{*}{\rotatebox[origin=c]{90}{MorGAN}}&BSIF+SVM~\cite{kannala2012bsif}&                      1.57 &1.42&1.30\\
                                &SIFT+SVM~ \cite{lowe1999object}&     43.50 &93.20&84.20\\
                                &LBP+SVM~ \cite{ojala1994performance}&  20.10 &52.70 & 32.30\\
                                &SURF+SVM~ \cite{bay2006surf}&  39.95 &80.00 & 72.60\\
                                 &RGB+DNN~ \cite{szegedy2017inception}&  2.44 &1.88 & 1.50\\
                                &\textbf{Ours}&                                 {\bf0.00} &{\bf0.00}&{\bf0.00}\\

\hline

\end{tabular}
\end{center}

\label{table:results_cross}
\end{table}

\section{Experimental Setup}
\subsection{Datasets}
In this study, three different morphed image datasets are used that are, the VISAPP17\cite{makrushin2017automatic}, LMA \cite{damer2018morgan}, and MorGAN \cite{damer2018morgan}. The VISAPP17 dataset is generated using landmark-based face morphing attack, followed by splicing. In the landmark-based morphing pipeline locations of the corresponding landmarks in two bona fide subjects are averaged, and facial regions are divided using Delaunay triangulation before their alpha blending. LMA is a landmark-based morphed image dataset, and MorGAN dataset is generated using a generative model, GAN to be exact. Contrary to the landmark-based morphing attack, which captures geometry of underlying bona fide images, GAN-based morphing attacks synthesize morphed images after capturing the underlying distributions of bona fide facial images.

MTCNN \cite{zhang2016joint} is utilized for face detection and alignment. Face images are resized to $160\times160$ pixels. For each dataset, 50\% of the subjects are considered for training while the other 50\% are used for the test set. In addition, 15\% of the training set is selected during model optimization as the validation set. The train-test split is disjoint, with no overlapping bona fides, morphs, or bona fides contributing to morphs. In addition to the individual datasets, we combine the three datasets into a {\it universal dataset}. Regarding the universal dataset, the training set includes 1631 bona fide, and 1183 morphed samples. The validation set contains 462 bona fide, and 167 morphed subjects. In addition, the test set is composed of 1631 bona fide, and 1183 morphed images.

\subsection{Training Setup}
For the backbone of our attention-based morph detector, we employ Inception-ResNet-v1 \cite{szegedy2017inception} , which harnesses the residual skips \cite{he2016deep}, as well as the revised version of the Inception network \cite{szegedy2016rethinking}. We add three attention modules to the network at $\mathcal{L}_1$ = {\it "conv2d\_4b"}, $\mathcal{L}_2$ = {\it "mixed\_6a"}, and $\mathcal{L}_3$ = {\it "mixed\_7a"}. Since the number of channels in the resulting feature vectors related to the three convolutional layers are not 512-D, we project the feature vectors to new vectors where number of channels are 512. The projection in each convolutional layer is achieved using the $1\times 1$ convolutional filters, where 512 kernels with the size of $1 \times 1$ are employed. The Adam optimizer updates the weights of our DNN accelerated using two 12 GB TITAN X (Pascal) GPUs. Batch size of $8$ is considered for training.

\begin{table}[t]
\caption[Table caption text]{Performance of single morph detection: D-EER\%, BPCER@APCER=5\%, and BPCER@APCER=10\%.} 
\small
\begin{center}
\addtolength{\tabcolsep}{-0pt}
\begin{tabular}{lccccc}
\hline
Train&Test&Algorithm&D-EER&5\%&10\%\\ 
\hline
\multirow{24}{*}{\rotatebox[origin=c]{90}{Universal(VISAPP17+LMA+MorGAN)}}&\multirow{6}{*}{\rotatebox[origin=c]{90}{VISAPP17}}&BSIF+SVM~ \cite{kannala2012bsif}&                      35.00 &67.20&59.00\\
                                &&SIFT+SVM~ \cite{lowe1999object}&     27.00 &83.20&70.90\\
                                &&LBP+SVM~ \cite{ojala1994performance}&  37.67 &72.50 & 59.50\\
                                &&SURF+SVM~ \cite{bay2006surf}&  31.00 &79.40 & 70.10\\
                                &&RGB+DNN~ \cite{szegedy2017inception}&  0.00 &0.00 & 0.00\\
                                &&\textbf{Ours}&                                 {\bf0.00} &{\bf0.00}&{\bf0.00}\\\cline{2-6}
                               
&\multirow{6}{*}{\rotatebox[origin=c]{90}{LMA}}&BSIF+SVM~ \cite{kannala2012bsif}&                      30.00 &70.42&57.60\\
                                &&SIFT+SVM~ \cite{lowe1999object}&     28.31 &67.70&50.00\\
                                &&LBP+SVM~ \cite{ojala1994performance}&  29.00 &61.50 & 51.20\\
                                &&SURF+SVM~ \cite{bay2006surf}&  33.40 &74.50 & 62.70\\
                                &&RGB+DNN~ \cite{szegedy2017inception}&  7.80 &13.00 & 6.10\\
                                &&\textbf{Ours}&                                 {\bf8.11} &{\bf14.21}&{\bf6.83}\\\cline{2-6}

&\multirow{6}{*}{\rotatebox[origin=c]{90}{MorGAN}}&BSIF+SVM~\cite{kannala2012bsif}&                      28.80 &62.42&45.70\\
                                &&SIFT+SVM~ \cite{lowe1999object}&     47.60 &92.30&88.60\\
                                &&LBP+SVM~ \cite{ojala1994performance}&  31.20 &62.00 & 55.60\\
                                &&SURF+SVM~ \cite{bay2006surf}&  38.67 &76.00 & 70.00\\
                                &&RGB+DNN~ \cite{szegedy2017inception}&  4.69 &4.70 & 2.74\\
                                &&\textbf{Ours}&                                 {\bf2.59} &{\bf1.50}&{\bf0.89}\\\cline{2-6}

&\multirow{6}{*}{\rotatebox[origin=c]{90}{Universal}}&BSIF+SVM~ \cite{kannala2012bsif}&                      23.74 &51.42&38.67\\
                                &&SIFT+SVM~ \cite{lowe1999object}&     37.21 &87.45&76.71\\
                                &&LBP+SVM~ \cite{ojala1994performance}&  38.80 &91.36 & 83.40\\
                                &&SURF+SVM~ \cite{bay2006surf}&  36.00 &75.50 & 65.76\\
                                &&RGB+DNN~ \cite{szegedy2017inception}&  5.57 &6.08 & 3.00\\
                                &&\textbf{Ours}&                                 {\bf6.42} &{\bf7.58}&{\bf3.46}\\\cline{2-6}

\end{tabular}
\end{center}
 
\label{table:results_cross}
\end{table}

\subsection{Performance of the Attention-based Morph Detector}
Standard quantitative measures are used to evaluate the effectiveness of our proposed method. The first measure is Attack Presentation Classification Error Rate (APCER), which is the percentage of morphed images that are classified as bona fide. The second measure is Bona Fide Presentation Classification Error Rate (BPCER), which represents the percentage of bona fide samples that are classified as morphed. If we label the morphed class as positive and the bona fide class as negative , APCER, and BPCER are equivalent to false negative rate and false positive rate, respectively. Detection error trade-off (DET) curves represent performance of our attention-based DNN. D-EER stands for the Detection Equal Error Rate, where APCER equals BPCER. BPCER5 represents BPCER rate for APCER=5\%, and BPCER10 represents BPCER rate for APCER=10\%.\\
We train our attention-based DNN using the three datasets, that are the VISAPP17, LMA, and MorGAN. Table 1 delineates the performance of the baseline methods, as well as our attention-based morph detector for the three datasets. In addition, Fig. 3 depicts the detection error trade-off (DET) curves for the three datasets.\\
Moreover, we scrutinize the scenario where all three datasets are combined, which was coined the universal dataset. Therefore, we train our network using the training portion of the universal dataset, and test set comes from all individual datasets, as well as the testing portion of the universal dataset. The performance of our attention-based morph detector when trained on the universal dataset is summarised in Table 2, and Fig. 4 depicts the DET curves when the attention-based DNN is trained using the universal dataset. Our attention-based morph detector can detect morphed samples in the VISAPP17 and MorGAN datasets accurately when the network is trained on each dataset.

\begin{figure}[t]
\begin{center}
%\fbox{\rule{0pt}{2in}
\includegraphics[width=.9\linewidth]{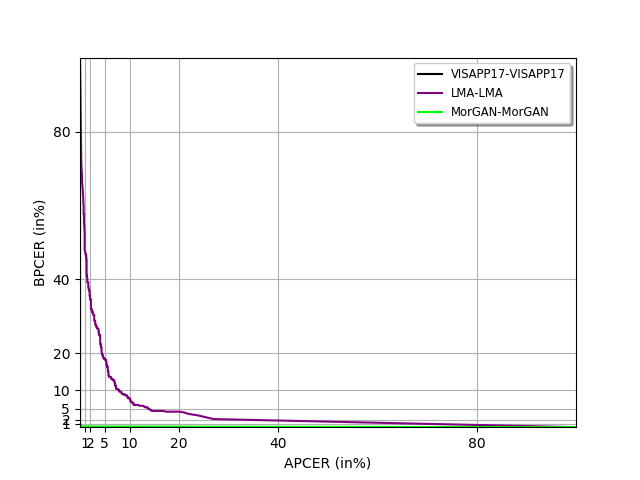} 
\end{center}
   \caption{DET curves when our attention-based morph detector is trained using the individual datasets.}
\end{figure}

\subsection{Estimated Attention Maps}
The estimated attention maps, resulting from the three attentions modules are shown in Fig. 5. It is worth mentioning that the heatmaps of the resulting attention maps are applied to the image for visualization purpose. The first row, shows the bona fide image, and its corresponding attention maps from the three different convolutional layers. The second row is related to the attention maps of the morphed image. Comparing the attention map of the $\mathcal{L}_1$ for the bona fide image with that of the morphed image reveals that the morphed images has more attended areas that is caused by the morphing attack pipeline, which comprises landmark manipulation for this image, coming from the VISAPP17 dataset. Given the attention maps of the $\mathcal{L}_2$ and $\mathcal{L}_3$, there are salient impacted regions in the feature maps of the morphed images, while there is no obvious attentive regions in the bona fide image.

\begin{figure}[t]
\begin{center}
%\fbox{\rule{0pt}{2in}
\includegraphics[width=.9\linewidth]{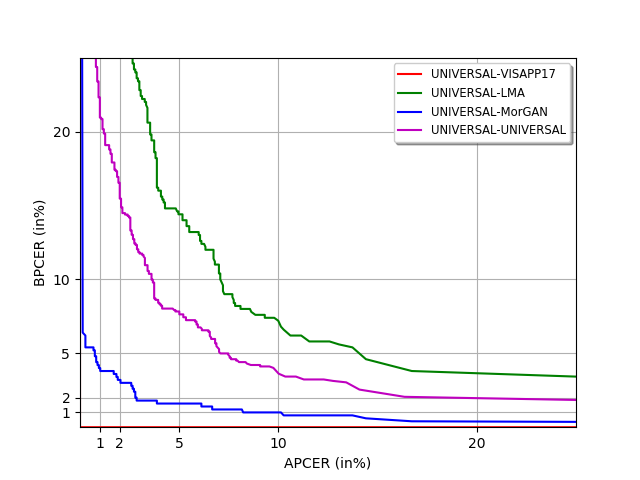} 
\end{center}
   \caption{DET curves when our attention-based morph detector is trained using the training portion of the universal dataset.}
\end{figure}

\begin{figure}[t]
\begin{center}
%\fbox{\rule{0pt}{2in}
\includegraphics[width=.9\linewidth]{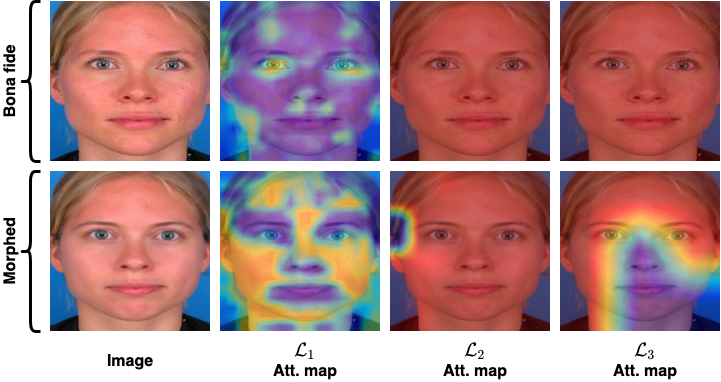} 
\end{center}
   \caption{Estimated attention maps for a bona fide and the corresponding morphed image obtained from the three attention modules.}
\end{figure}

\subsection{Ablation Study}
In this section, we delve into the effect of the attention modules on the performance of our attention-based morph detector. To this end, we compare the performance of the morph detector when the number of attention modules are one, two, or three. We have already studied the case where number of attention modules are three in the section 4.3. We plot the DET curves for the individual datasets for every number of attention modules. Table 3 delineates the performance of our morph detector when trained on the individual datasets for the following cases: 1- One attention module placed at $\mathcal{L}_3$ = {\it "mixed\_7a"}, 2- two attention modules placed at $\mathcal{L}_3$ = {\it "mixed\_7a"} and $\mathcal{L}_2$ = {\it "mixed\_6a"}, 3- three attention modules at $\mathcal{L}_3$ = {\it "mixed\_7a"}, $\mathcal{L}_2$ = {\it "mixed\_6a"}, and $\mathcal{L}_1$ = {\it "conv2d\_4b"}. Also, Table 4 summarizes the performance of our morph detector when trained on the universal dataset for the above-mentioned cases.

Fig. 6 depicts the performance of our morph detector using the two attention modules when our morph detector is trained using the individual datasets. Fig. 7 shows the performance of our morph detector using the two attention modules when our DNN is trained using the universal dataset. Moreover, Fig. 8 displays the performance of our attention-based morph detector with the one attention module that is trained on the individual datasets, and Fig. 9 displays the performance of our attention-based morph detector using the one attention module when the network is trained on the universal dataset. It is evident from Table 3 that the most accurate morph detection for the LMA dataset is achieved when there are three attention modules in our proposed network. Concerning Table 4, the attention-based DNN with three attention modules outperforms the network which has either one or two attention modules.

\begin{table}[t]
\caption[Table caption text]{Performance of single morph detection for the different number of attention-modules: D-EER\%, BPCER@APCER=5\%, and BPCER@APCER=10\%.} 
\small
\begin{center}
\addtolength{\tabcolsep}{-0pt}
\begin{tabular}{llccc}
\hline
Dataset&Att. Layers&D-EER&5\%&10\%\\ 
\hline
\multirow{3}{*}{{VISAPP17}}&$\mathcal{L}_3$&                      00.00 &00.00&00.00\\
                                &$\mathcal{L}_2$+$\mathcal{L}_3$&00.00 &00.00&00.00\\
                                &$\mathcal{L}_1$+$\mathcal{L}_2$+$\mathcal{L}_3$&  00.00 &00.00&00.00\\

% &\multirow{5}{*}{\rotatebox[origin=c]{90}{Universal}}&BSIF+SVM~ \cite{kannala2012bsif}&                      23.74 &51.42&38.67\\
%                                 &&SIFT+SVM~ \cite{lowe1999object}&     37.21 &87.45&76.71\\
%                                 &&LBP+SVM~ \cite{ojala1994performance}&  38.80 &91.36 & 83.40\\
%                                 &&SURF+SVM~ \cite{bay2006surf}&  36.00 &75.50 & 65.76\\
%                                 &&Ours&                                 {\bf3.55} &{\bf2.69}&{\bf1.27}\\\cline{2-6}                                       
\hline

\multirow{3}{*}{{LMA}}&$\mathcal{L}_3$&                      12.45 &21.23&15.18\\
                                &$\mathcal{L}_2$+$\mathcal{L}_3$&12.12 &23.58&17.21\\
                                &$\mathcal{L}_1$+$\mathcal{L}_2$+$\mathcal{L}_3$&  8.71 &17.86 & 6.52\\

\hline

\multirow{3}{*}{{MorGAN}}&$\mathcal{L}_3$&                      00.00 &00.00&00.00\\
                                &$\mathcal{L}_2$+$\mathcal{L}_3$&00.00 &00.00&00.00\\
                                &$\mathcal{L}_1$+$\mathcal{L}_2$+$\mathcal{L}_3$&  00.00 &00.00&00.00\\
\hline

\end{tabular}
\end{center}

\label{table:results_cross}
\end{table}

\begin{table}[t]
\caption[Table caption text]{Performance of the universal training set single morph detection for different number of attention-modules: D-EER\%, BPCER@APCER=5\%, and BPCER@APCER=10\%.}
\small
\begin{center}
\addtolength{\tabcolsep}{-0pt}
\begin{tabular}{cllccc}
\hline
Train&Test&Att. Layers&D-EER&5\%&10\%\\ 
\hline
\multirow{12}{*}{\rotatebox[origin=c]{90}{Universal}}&\multirow{3}{*}{{VISAPP17}}

&$\mathcal{L}_3$&00.00 &00.00&00.00\\
&&$\mathcal{L}_2$+$\mathcal{L}_3$&00.00 &00.00&00.00\\
&&$\mathcal{L}_1$+$\mathcal{L}_2$+$\mathcal{L}_3$&  00.00 &00.00&00.00\\
\cline{2-6}
                               
&\multirow{3}{*}{{LMA}}
&$\mathcal{L}_3$&14.37 &27.23&16.54\\
&&$\mathcal{L}_2$+$\mathcal{L}_3$&13.24 &35.36&18.61\\
&&$\mathcal{L}_1$+$\mathcal{L}_2$+$\mathcal{L}_3$&  8.11 &14.21 & 6.83\\\cline{2-6}

&\multirow{3}{*}{{MorGAN}}
&$\mathcal{L}_3$& 7.21 &6.31&5.02\\
&&$\mathcal{L}_2$+$\mathcal{L}_3$&7.14 &7.86&4.91\\
&&$\mathcal{L}_1$+$\mathcal{L}_2$+$\mathcal{L}_3$&  2.59 &1.50 & 0.89\\\cline{2-6}

&\multirow{3}{*}{{Universal}}
&$\mathcal{L}_3$&8.91 &12.21&8.27\\
&&$\mathcal{L}_2$+$\mathcal{L}_3$&9.95 &12.23&8.93\\
&&$\mathcal{L}_1$+$\mathcal{L}_2$+$\mathcal{L}_3$&  6.42 &7.58 & 3.46\\\cline{2-6}

\end{tabular}
\end{center}
  
\label{table:results_cross}
\end{table}

\begin{figure}[t]
\begin{center}
%\fbox{\rule{0pt}{2in}
\includegraphics[width=.9\linewidth]{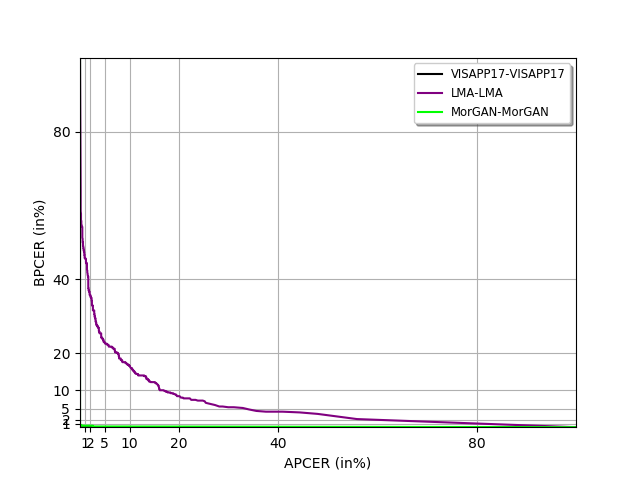} 
\end{center}
   \caption{DET curves for the individual datasets for two attention modules.}
\end{figure}

\begin{figure}[t]
\begin{center}
%\fbox{\rule{0pt}{2in}
\includegraphics[width=.9\linewidth]{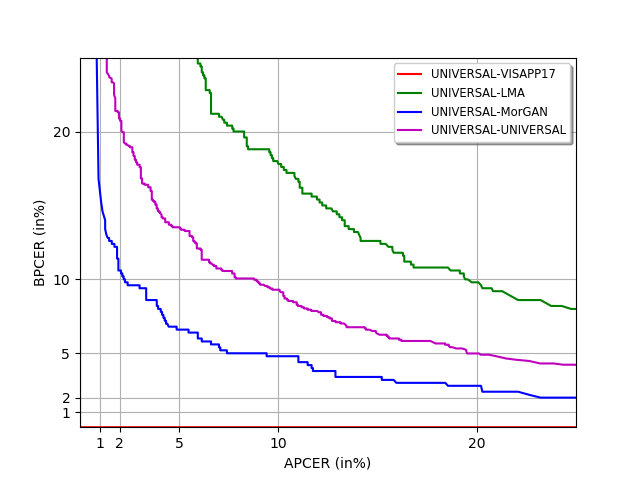} 
\end{center}
   \caption{DET curves for the universal datasets for two attention modules.}
\end{figure}

\begin{figure}[t]
\begin{center}
%\fbox{\rule{0pt}{2in}
\includegraphics[width=.9\linewidth]{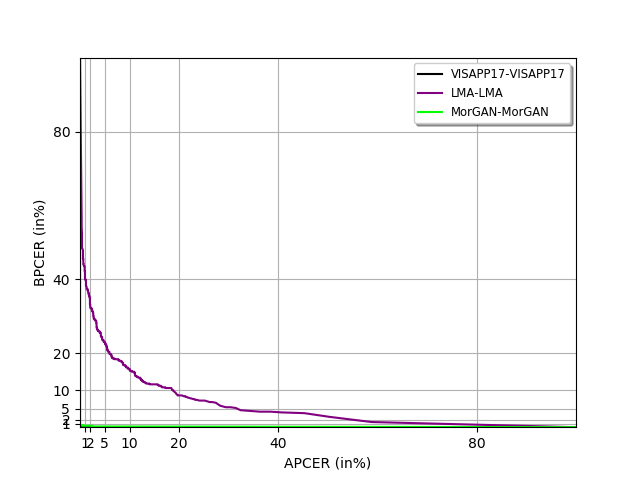} 
\end{center}
   \caption{DET curves for the individual datasets for one attention module.}
\end{figure}

\begin{figure}[t]
\begin{center}
%\fbox{\rule{0pt}{2in}
\includegraphics[width=.9\linewidth]{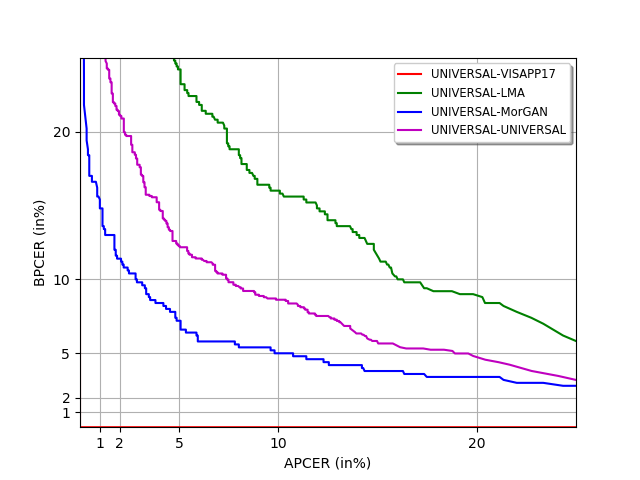} 
\end{center}
   \caption{DET curves for the universal datasets for one attention module.}
\end{figure}

\section{Conclusion}
In this paper, we study the application of attention mechanism for detecting morphed images. More importantly, our attention-based model is adapted to a wavelet-based Inception-ResNet-v1, where all input images are decomposed into 48 wavelet sub-bands. The three integrated attention modules can emphasize the artifacts stem from the morphing attack, leading to detecting morphed images accurately. Most importantly, our attention-based morph detector can detect morphed images in the VISAPP17 and MorGAN datasets accurately. Displayed attention maps substantiates the effectiveness our algorithm in detecting morphed images, because morphed images have substantial attentive pixels compared to bona fide images. Finally, our ablation study proves the superior performance of our attention-based morph detector that uses three attention modules in comparison to a network that has either one or two attention modules. 

{\small
\bibliographystyle{ieee}
\bibliography{submission_example}
}

\end{document}